\documentclass{article}
\usepackage{microtype}
\usepackage{amsfonts}
\usepackage{graphicx}
\usepackage{subfigure}
\usepackage{booktabs} % for professional tables
\usepackage{hyperref}
\usepackage{enumitem,xspace}
\usepackage{amsmath}
\usepackage{amssymb}
\usepackage{multirow}

\usepackage[accepted]{icml2022}

\usepackage{tabularx}

\newcommand{\proposal}{\texttt{FedSIM}\xspace}

\icmltitlerunning{Personalized Federated Learning with Server-Side Information}

\begin{document}

\twocolumn[
\icmltitle{Personalized Federated Learning with Server-Side Information}

\begin{icmlauthorlist}
\icmlauthor{Jaehun Song}{snu}
\icmlauthor{Min-hwan Oh}{snu}
\icmlauthor{Hyung-Sin Kim}{snu}\\
Graduate School of Data Science, Seoul National University, Seoul, South Korea\\
\{steve2972\}, \{minoh\}, \{hyungkim\} @snu.ac.kr
\end{icmlauthorlist}
\icmlaffiliation{snu}{snu}%{}
\icmlkeywords{Machine Learning, Federated Learning, Meta Learning, Personalization}

\vskip 0.3in
]

% Input Sections of Research Paper
\begin{abstract}
Personalized Federated Learning (FL) is an emerging research field in FL that learns an easily adaptable global model in the presence of data heterogeneity among clients.
% [Main problem]
However, one of the main challenges for personalized FL is the heavy reliance on clients' computing resources to calculate higher-order gradients since client data is segregated from the server to ensure privacy.
% [Problem formulation]
To resolve this, we focus on a problem setting where the server may possess its own data independent of clients' data -- a prevalent problem setting in various applications, yet relatively unexplored in existing literature.
% [Problem Solution]
Specifically, we propose \proposal, a new method for personalized FL that actively utilizes such \textit{server data} to improve meta-gradient calculation in the server for increased personalization performance.
% [Experimental Results]
Experimentally, we demonstrate through various benchmarks and ablations that \proposal is superior to existing methods in terms of accuracy, more computationally efficient by calculating the full meta-gradients in the server, and converges up to 34.2\% faster. 

\end{abstract}

\section{Introduction}

%\MHO{(MHO: Perhaps, we can remove this entire paragraph and start from the next paragraph with a little modification.)}
%Mobile devices, such as smartphones and wearable devices, are increasingly becoming the primary computing device for most people~\cite{mobilephonesprimarycomputing}.
%With their improving computing and sensing capabilities, these devices have amassed a vast amount of personal data that enrich user experience in various applications~\cite{healthcareai, dhar2020ondevice, gboard, sym12040499}.
%However,  with privacy becoming an increasingly important issue in the digital era, utilizing personal data to train models in a central server holds substantial risks and responsibilities.

%

Federated Learning (FL) has drawn significant attention from the research community in recent years due to its potential for privacy-centric machine learning in distributed learning environments. However, one challenge of FL that remains prevalent today is diverse, \textit{non-i.i.d.} data distributions among clients, which limits a single global model from delivering optimal performance on each client's task.

%Thus, due to its potential for privacy-centric machine learning, Federated Learning (FL) has drawn significant attention from the research community, providing an alternative approach for distributed training without sharing personal data with a centralized server~\cite{mcmahan2017communicationefficient}.

%Given that FL suffers performance degradation with \textit{non-i.i.d.} client data, 
One of the
recent research directions
% attempts 
that address this issue is
% by introducing 
\textit{personalized federated learning}, a personalized variant of FL based on techniques used in 
optimization-based
meta-learning such as in the Model-Agnostic Meta-Learning (MAML) framework \cite{finn2017modelagnostic}. 
The goal of personalized FL is to create an \textit{adaptable} global model parametrized by $\theta$ in a federated environment such that $\theta$ can easily be fine-tuned to each client's %\MHO{(MHO: should we use ``user'' or ``client''?)} 
% personal 
individual task  with a small number of gradient steps. 
%\HSK{Here we call the global model $\theta$ \textit{adaptable} if once $\theta$ is well trained and deployed at users, each user easily fine-tunes $\theta$ with a small number of gradient steps for its personalized model.}
To achieve this goal, personalized FL optimizes 
%
% \begin{equation}
% \label{eq: perfl_opt_problem}
% \min_{\theta \in \mathbb{R}^d}F(\theta) := \frac{1}{n}\sum_{i=1}^n f_i(\phi_i)
% \end{equation}

\begin{equation}
\label{eq: perfl_opt_problem}
\min_{\theta \in \mathbb{R}^d}F(\theta) := \frac{1}{n}\sum_{i=1}^n f_i(\theta - \eta\nabla f_i(\theta; \mathcal{D}_i))
\end{equation}
where $f_i : \mathbb{R}^d \rightarrow \mathbb{R}$ is the loss 
% and $\phi_i := \theta - \eta\nabla f_i(\theta; \mathcal{D}_i)$ are the optimized model parameters 
corresponding to client $i$, $\eta \geq 0$ is the step-size, and $\mathcal{D}_i$ is a batch of data for client $i$ which follows the distribution $p_i$ where each client's data is assumed to be \textit{heterogeneous}.
%After the adaptable model $\theta$ is deployed,
%(\MHO{MHO: I don't understand what ``the adaptable model $\theta$'' means...})
%users fine-tune $\theta$ with a small number of gradient steps for a personalized model. 

%However, a crucial caveat of meta-learning is the necessity to calculate second-degree meta-gradients. These gradients require not only data sampled from the client's data distribution such that $\mathcal{D}^q_i, \mathcal{D}^h_i \sim p(\mathcal{D}_i)$, but also additional computation to leverage this data into Hessians.
%This in turn becomes a major bottleneck during the implementation of meta-learning in resource-constrained mobile environments for personalized federated learning.

However, a major challenge of
% a crucial caveat of 
applying meta-learning to FL is the necessity 
to calculate second-degree meta-gradients~\cite{fallah2020personalized,chen2019federated}. 
These gradients require not only additional data sampled from each client's data distribution $p_i$, but also additional computation for clients to obtain the required Hessian matrix.
%Prior work tries to avoid this problem by disregarding second-degree meta gradients~\cite{jiang2019improving}, which reduces model accuracy.
%
Hence, computing the Hessian poses a significant bottleneck
% this becomes a major bottleneck during the implementation of meta-learning 
in resource-constrained environments for personalized FL.

To address this challenge,
% solve this problem, 
we aim to efficiently utilize computational resources available in the entire system,  acknowledging the \textit{disparity of the computational overhead} between clients and the central server in the existing methods.
% first focus on  the \textit{disparity of the computational overhead} between clients and the central server. 
Previous methods in personalized FL postulate that the central server needs only to aggregate and average the optimized client weights to update the global model. Here, the resourceful server is idle for most of the training process while the resource-constrained clients are busy optimizing their local models.

In this paper, we consider a 
variant of the personalized FL problem where the server contains its own data.
We denote this problem setting as
% new scenario
\textit{Personalized FL with Server Data}. 
%Although previous research work around this problem by either disregarding higher-degree gradients~\cite{jiang2019improving} or calculating the Hessians within the clients~\cite{fallah2020personalized,chen2019federated},  these methodologies either reduce model accuracy or increase the computational burden of clients respectively. We instead formulate a method to estimate the meta-gradients in the server by using \textit{server data}.
\textit{Server data} is defined as data 
% by which developers 
used to create and test a model in the server before initiating the FL process and can be available
% seen used 
in various application domains.
% scenarios.
For example, hospitals and healthcare providers
% that oversee federated learning to create models for predicting diseases usually 
may
first test the validity of models using their own records before implementing patient-wise predictions based on more privacy-sensitive individual records. 
In predictive text, an initial predictive model can be trained 
% Another use-case of server data is training an initial predictive 
% keyboard 
at the server with common phrases or words before implementing large-scale FL for each client's mobile device. 
In addition, an autonomous driving company gathers its own data in various road conditions to train a model, but can utilize FL to improve the model for each driver.  
However, most FL methods use such server data only for creating an initial model and \textit{disregard it during the FL process}.

\begin{figure}[t]
    \centering
    \includegraphics[width=\linewidth]{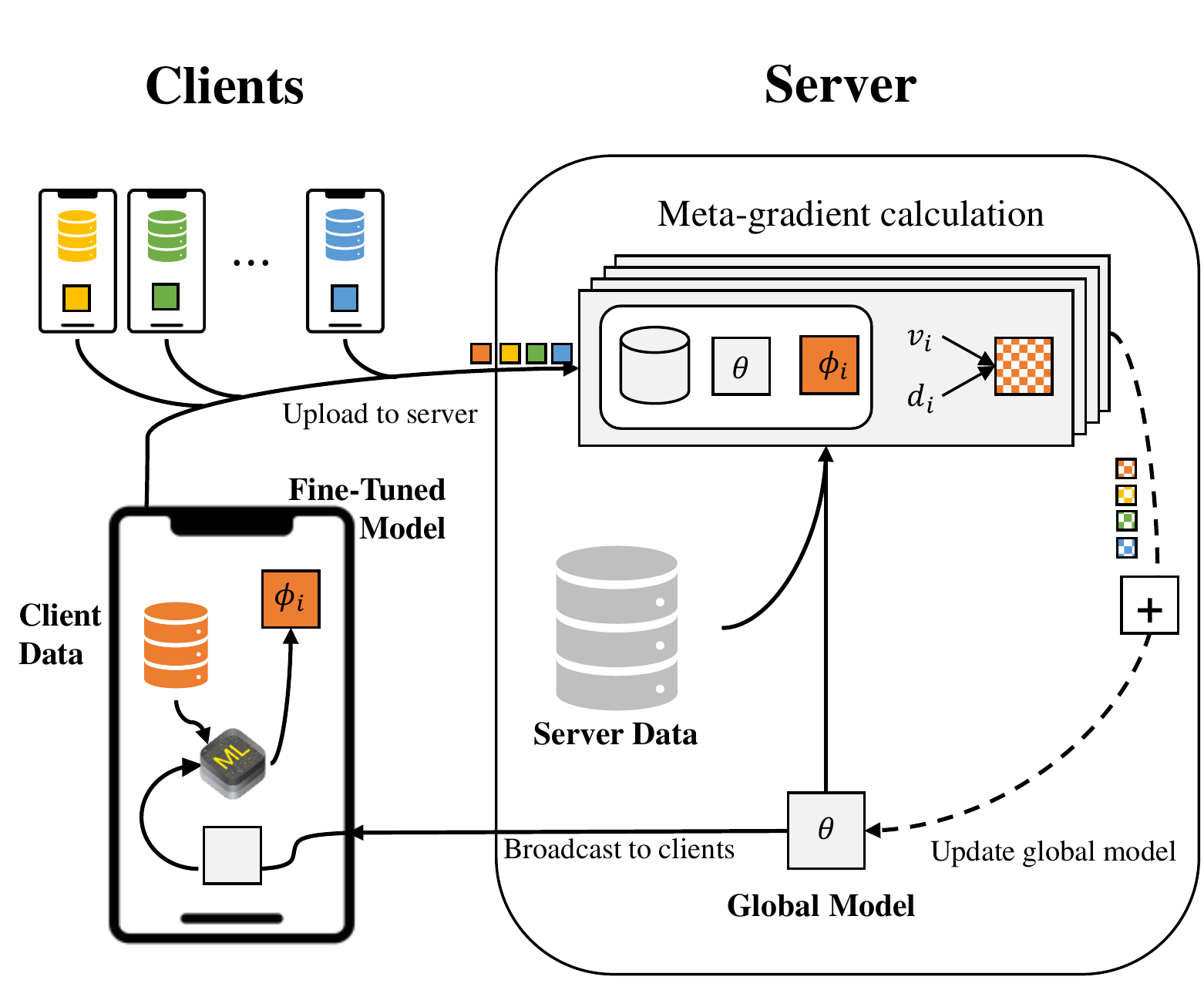}
    \vspace{-3ex}
    \caption{Overall architecture of \proposal. Clients focus solely on local optimization while the server calculates  meta-gradients for all clients using server data.}
    \label{fig:fedsim}
    \vspace{-3ex}
\end{figure}

To this end, we propose a new method to estimate the computationally heavy meta-gradients in the server by using \textit{server data}.
We propose that server data  can actively be utilized during the federated training process to augment model performance, as described in Figure~\ref{fig:fedsim}.
We summarize our main contributions as follows.
\vspace{-2ex}
\begin{itemize}[leftmargin=*]
   % \item %We design \proposal, a new framework for personalized FL that generates an adaptable global model without additional client computation compared to vanilla federated learning. 
    %
    \item We propose a novel method \proposal for
    % a practical problem of FL, namely 
    Personalized Federated Learning with Server Data. To our knowledge, \proposal is the first personalized FL method that
    % show that 
    efficiently utilizes
    the server's computational and information resources 
    % can be effectively utilized 
    to compute the estimates of full meta-gradients 
    % in the context of personalized FL 
    with no additional client computation compared to conventional FL.

    \item 
    % We design and test \textbf{three analytical components} that run in each round of \proposal: 
    The key components of our proposed method include
    (\romannumeral 1) a custom loss with $L_2$ regularization for local optimization, (\romannumeral 2) the approximation of first-order meta gradients for each client 
    by using the differences between personalized model parameters and global model parameters,
    % using its optimized local weights
    (\romannumeral 3) the approximation
    % calculation 
    of second-order meta gradients for each client using server data 
    without explicitly computing Hessian matrices.
    
    \item 
    % We empirically 
    The empirical evaluations
    demonstrate that \proposal 
    % successfully 
    effectively
    improves model performance \textbf{even when the server has a relatively small amount of data} compared to the entire dataset ($\le$5\%), or when the distribution of server data weakly represents that of non-i.i.d. data for each client.
    
    \item We show that \proposal \textbf{outperforms existing methods} in personalized FL. In standardized FL benchmarks proposed in~\cite{chaoyanghe2020fedml}, \proposal is up to 2.57\% more accurate and requires 34.2\% less communication rounds for convergence. 
\end{itemize}

\section{Related Work}

\subsection{Federated Learning}
Federated learning 
has rapidly evolved in various aspects \cite{Li_2020}, with both empirical analyses~\cite{bonawitz2019federated} and theoretical guarantees~\cite{hanzely2020lower} showing that FL models 
% trained in this manner 
exhibit similar performance to models trained in centralized data centers even when data does not leave clients. In particular, there have been several works on the various aspects of FL, including methods of reducing communication costs through quantization~\cite{amiri2020federated, sun2019communicationefficient} or adaptive gradient upload rounds~\cite{wang2019adaptive, amiri2020federated, ivkin2020communicationefficient}, and convergence analyses with well-defined lower bounds~\cite{hanzely2020lower, pathak2020fedsplit, wang2019adaptive}.

A fundamental problem of FL is accuracy degradation due to training the model with  non-i.i.d. data across clients~\cite{zhao2018federated}. This problem is significant because heterogeneous data distributions are common in practice~\cite{bonawitz2019federated} and thus investigated in a number of studies~\cite{haddadpour2019convergence, khaled2020tighter,li2019convergence} with solutions including normalized federated updates~\cite{wang2020tackling} and computing stochastic gradients in minibatches~\cite{woodworth2020minibatch}.

\subsection{Personalized Federated Learning}
Personalized Federated Learning is a personalized variant of federated learning that
% has also emerged in recent years that
aims to improve model performance in non-i.i.d. data settings. Examples include using Moreau envelopes~\cite{moreau_envelopes}, model interpolation~\cite{mansour2020approaches} and transfer learning-based personalization~\cite{fedper, chen2021fedhealth}.

In particular, \texttt{FedMeta}~\cite{chen2019federated} and Per-FedAvg~\cite{fallah2020personalized} consider building upon the Model-Agnostic Meta-Learning (MAML) formulation~\cite{finn2017modelagnostic}, and study the empirical and theoretical success of the framework in a federated environment. 
However, these approaches require the resource-constrained clients to locally execute full Hessian calculations, thereby significantly increasing client-side computation and memory overhead. 
A number of other works aim to decrease this computational bottleneck by disregarding second-order calculations~\cite{jiang2019improving}, inspired by first-order gradient-based meta learning approaches as in~\cite{reptile}, while sacrificing model performance.

In contrast, \proposal aims to calculate heavy meta gradients at the server using server data to mitigate accuracy degradation due to disregarding the Hessians without additional computational burden on clients. 

\section{Federated Learning with Server Information Meta-Learning (FedSIM)}
\label{sec:method}

\subsection{Problem: Personalized FL with Server Data}
In conventional FL, there are $n$ clients in a federated environment that tries to find a global model $\theta$ by optimizing the following problem:
\begin{equation}
\label{eq:fed_opt_problem}
\min_{\theta \in \mathbb{R}^d} f(\theta) := \frac{1}{n}\sum_{i=1}^n f_i(\theta)
\end{equation}
where $f_i: \mathbb{R}^d \rightarrow \mathbb{R}$ ($i = 1, ..., n$) denotes the expected loss over the data distribution of client $i$ such that 
\begin{equation}
\label{eq:loss}
f_i(\theta) = \mathbb{E}_{\mathcal{D}_i \sim p_i} [\ell_i (\theta; \mathcal{D}_i)]
\end{equation}
where $\mathcal{D}_i$ is a random data sample drawn from client $i$'s data distribution $p_i$ and $\ell_i(\theta; \mathcal{D}_i)$ is the loss corresponding with this data sample w.r.t. a global model parameter $\theta$.
% \MHO{(I suggest we use a different notation for $\ell$..., e.g., $\ell_i(\theta; \mathcal{D}_i)$ for the loss, maybe.)}

In contrast to Eq. (\ref{eq:fed_opt_problem}), we learn
% generate 
an \textit{adaptable} global model $\theta$ in a federated environment, by formulating and solving a  bi-level (server- and client-side) problem  defined as  
\begin{equation}
\label{eq: perfl_opt_problem_server}
\min_{\theta \in \mathbb{R}^d} F(\theta) := \frac{1}{n}\sum_{i=1}^n F_i(\theta)
\end{equation}
\begin{equation}
\label{eq: perfl_opt_problem_client}
F_i(\theta) = \min_{\phi_i \in \mathbb{R}^d} \tilde{f}_i(\phi_i) := f_i(\phi_i) + \frac{\lambda}{2}||\phi_{i}-\theta||^2_2 
\end{equation}
where $\phi_i$  denotes the \textit{personalized model} of client $i$ and $\lambda$ is a regularization parameter.
% \MHO{(Are you using $|| \cdot ||$ as $\ell_2$ norm? Then need to be specific.)}. 
% a scalar hyperparameter that denotes regularization strength.
Note that instead of the conventional loss in Eq.\eqref{eq:loss}, we define a different loss function $\tilde{f}_i(\phi_i)$ that includes an $L_2$ regularization term such that its gradient becomes  
\begin{equation}
\label{eq:lossgradients}
\nabla_{\phi}\tilde{f}_i(\phi_{i})  
 = \nabla_{\phi} f_i(\phi_{i}) + \lambda(\phi_{i} - \theta)  
\end{equation}
%
%\begin{equation}
%\label{eq:lossfuction}
%\ell_i'(\phi_i, \mathcal{D}_i) = \ell_i(\phi_i, %\mathcal{D}_i) + \frac{\lambda}{2}||\phi_{i}-\theta||^2
%\end{equation}
%
%where $\lambda$ is a scalar hyperparameter that denotes regularization strength. 
This custom loss is in accordance with ideas from \cite{li2020fedopt,moreau_envelopes} such that the personalized parameters $\phi_i$ are encouraged to tend towards the global parameters $\theta$,
% \MHO{(Aren't we overloading the term $\theta$? Perhaps, we can write ``initial'' parameter differently.)}
which improves convergence of the global model in non-i.i.d. data settings.

Lastly, we make a practical assumption that is not included in previous work on personalized FL; the server has \textit{its own data} with distribution $p_s$ independent of clients' data. Here, we also assume that the proportion of server data is small compared to the entire dataset. 

%Thus, we aim to solve for the optimization problem (\ref{eq: %perfl_opt_problem}) where
%
%\begin{equation}
%\label{eq:loss_optimization}
%f_i(\theta) = \min_{\phi_i \in \mathbb{R}^d} \left\{{f_i(\phi_i) + %\frac{\lambda}{2}}||\phi_{i}-\theta||^2\right\}
%\end{equation}
%\vspace{-2ex}

%%%%%%%%%%%%%%%%%%%%%%%%%%%%%%%%%%%%%%%%%%%%%%%%%%%%%%%%%%%%
%%%%%%%%%%%%%%%%%%%%%%%%%%%%%%%%%%%%%%%%%%%%%%%%%%%%%%%%%%%%
\subsection{\texttt{FedSIM}: A FL Framework for Server Utilization}
\textbf{Fed}erated Learning with \textbf{S}erver-Side \textbf{I}nformation \textbf{M}eta-Learning (\proposal) is a personalized FL  framework that aims to (\romannumeral 1) ensure client data privacy with (\romannumeral 2) minimal additional computation/communication overhead in clients compared to \texttt{FedAvg} in order to (\romannumeral 3) produce high-quality meta gradients. 
In this section, we present the \proposal algorithm to solve for Eq.\eqref{eq: perfl_opt_problem_server} in the context of Eq.\eqref{eq: perfl_opt_problem_client}.

In vanilla FL (e.g., \texttt{FedAvg}~in \citealt{mcmahan2017communicationefficient}), a centralized server computes
% generates 
% \MHO{(perhaps better not to use the term ``generate'' here since it can confuse the reader with generative process. This term appears multiple times across the paper. Isn't it simply ``compute"?)}
a global model by averaging models from decentralized devices. At each round $t$, the server samples a client subset $S_t$ of size $m$ to optimize the global model $\theta_{t-1}$. 
Each client $i\in S_t$ updates $\theta_{t-1}$ with its private data $\mathcal{D}_i \sim p_i$ using gradient decent for $E$ epochs and uploads the optimized model $\phi_i$ back to the server. Finally, the server updates the global model to $\theta_t$ by averaging $\phi_i$ received from $S_t$. It is important to note that $\mathcal{D}_i$ was never shared between the clients nor with the server.

%%%%%%%%%%%%%%%%%%%%%%%%%%%%%%%%%%%
%%%%%%%%%%%%%%%%%%%%%%%%%%%%%%%%%%%
%\textbf{Differentiation.}
%
\textbf{The main contribution of this work }comes from allocating the calculation of $\nabla_\theta F_i(\theta)$ to optimize Eq.\eqref{eq: perfl_opt_problem_server} between the clients and the server such that the server can calculate meta-gradients for multiple tasks without sharing data. To this end, \proposal follows the same principles as FL, but with additional computation for meta-gradients in the server to learn an easily adaptable global model.

\begin{algorithm}[ht]
	\caption{\proposal: Client-Side}
    \label{algo:proposal_client}
\begin{algorithmic}
\REQUIRE{Step size $\alpha$, regularization strength $\lambda$, client data distribution $p_i$}
\FUNCTION {ClientUpdate$(i, \theta)$\textbf{:} // \textit{Run on client} $k$}
\STATE $\phi_{i, 0} \gets \theta$
\FOR{each local epoch $e$ from $1$ to $E$}
\STATE Sample a mini-batch $\mathcal{D}_i$ from distribution $p_i$
\STATE Calculate $\phi_{i, e}$ using $\mathcal{D}_i$ with Eq.\eqref{eq:perfed_meta_gradients}
\ENDFOR
\STATE Return $\phi_{i, E}$ to server
\ENDFUNCTION
\end{algorithmic}
\end{algorithm}

%%%%%%%%%%%%%%%%%%%%%%%%%%%%%%%%%%%%%%%%%%%%%%%%%%%%%%%%%%%%
%%%%%%%%%%%%%%%%%%%%%%%%%%%%%%%%%%%%%%%%%%%%%%%%%%%%%%%%%%%%
\textbf{Client-side algorithm.} 
The client's main goal is to learn a personalized model $\phi_i$ by calculating local gradient updates at each local epoch $e$ as
\begin{equation}
\label{eq:perfed_meta_gradients}
\phi_{i,e} = \phi_{i,e-1} - \alpha \nabla_\phi\tilde{f}_i(\phi_{i,e-1})  
\end{equation}
To calculate $\nabla_\phi\tilde{f}_i(\phi_{i,e-1})$ in practice, we use an unbiased estimate $\nabla_{\phi} \ell_i(\phi_{i,e-1}; \mathcal{D}_i)$ by sampling a mini-batch of data $\mathcal{D}_i$ from distribution $p_i$. This process is illustrated in Algorithm \ref{algo:proposal_client}.

%While the client behavior is similar to that in \texttt{FedProx}, the key difference in \proposal is enabling \textit{personalized} FL without additional burden on clients by utilizing the server and its own data for meta-gradient calculation. 

\begin{algorithm}[ht]
    \caption{\proposal: Server-Side}
    \label{algo:proposal_server}
\begin{algorithmic}
    \REQUIRE{Step size $\beta$, $\delta$, server data distribution $p_s$}
    \STATE Initialize $\theta_0$
    \FOR{each round $t=1,2,...$}
    	\STATE Sample a mini-batch  $\mathcal{D}_s$ from distribution $p_s$
    	\STATE $S_t \gets $ Random subset of $m$ clients ($1 \le m \le n$)
    	\FOR{each client $i \in S_t$ \textbf{in parallel}}
        		\STATE $\phi_i \gets$ ClientUpdate($i$, $\theta_{t-1}$) (Algorithm~\ref{algo:proposal_client})
        		\STATE Calculate $v_i = \nabla_\phi f_i(\phi_i)$ with Eq.\eqref{eq:reptile}
        		\STATE Calculate $d_i = \nabla^2_\phi f_i(\phi_i)v_i$ using $\mathcal{D}_s$ with Eq.\eqref{eq:secondorderhessian}
        		\STATE Calculate meta-gradient $\nabla_\theta F_i (\theta_{t-1}) = v_i - \delta d_i$
       	 		\STATE Update $\tilde{\phi}_i \gets \phi_i - \beta \nabla_\theta F_i (\theta_{t-1})$
    	\ENDFOR
   	 	\STATE $\theta_{t} \gets \frac{1}{m}\sum_{i \in S_t}\tilde{\phi}_i$
    \ENDFOR
\end{algorithmic}
\end{algorithm}

%%%%%%%%%%%%%%%%%%%%%%%%%%%%%%%%%%%%%%%%%%%%%%%%%%%%%%%%%%%%
%%%%%%%%%%%%%%%%%%%%%%%%%%%%%%%%%%%%%%%%%%%%%%%%%%%%%%%%%%%%
\textbf{Server-side algorithm.} 
The server then attempts  to optimize Eq.\eqref{eq: perfl_opt_problem_server} for multiple communication rounds in Algorithm \ref{algo:proposal_server}. In each round $t$, the central server (\romannumeral 1) samples $m$ clients, (\romannumeral 2) calculates meta-gradients $\nabla_\theta F_i(\theta_{t-1})$ for each of these clients using local model $\phi_i$ and server data $\mathcal{D}_s$ and (\romannumeral 3) updates the global model from $\theta_{t-1}$ to $\theta_t$ using these meta-gradients.

As shown in~\cite{imaml}, the gradient of Eq.\eqref{eq: perfl_opt_problem_client} w.r.t. $\theta$ with the local loss function $\tilde{f}_i(\phi_i)$ can be written as
\begin{eqnarray}
\nabla_\theta F_i(\theta_{t-1}) = \left(\textbf{I}+\frac{1}{\lambda} \nabla^2_\phi f_i(\phi_i)\right)^{-1}\nabla_\phi f_i(\phi_i)
\label{eq:metagradient}
\end{eqnarray}
Note that $\nabla_\theta F_i(\theta_{t-1})$ is not dependent on the original meta model $\theta_{t-1}$, while corresponding with the personalized model $\phi_i$. This characteristic comes from the regularization term in Eq.~\eqref{eq: perfl_opt_problem_client}~\cite{imaml}. Since the meta-gradient $\nabla_\theta F_i(\theta_{t-1})$ is decoupled from $\theta_{t-1}$, the server approximates the meta-gradient without requiring a history of client $i$'s local updates. This allows clients to utilize multi-step gradient decent for local optimization.

We can see from Eq.\eqref{eq:metagradient} that the calculation of $\nabla_\theta F_i(\theta_{t-1})$ requires two terms: 
\begin{enumerate}
\item A first-order gradient $v_i = \nabla_\phi f_i(\phi_i)$
\item A Hessian-vector product $d_i = \nabla^2_\phi f(\phi_i) v_i$
\end{enumerate}
Unlike previous meta-learning approaches to personalized FL, \textbf{we propose to calculate both} $v_i$ and $d_i$ \textbf{using the server}, without requiring additional information or computation from clients.

%%%%%%%%%%%%%%%%%%%%%%%%%%%%%%%%%%%%%%%%%%%%%%%%%%%%%%%%%%%%
%%%%%%%%%%%%%%%%%%%%%%%%%%%%%%%%%%%%%%%%%%%%%%%%%%%%%%%%%%%%
% First-order meta gradient
\textbf{First-order meta-gradient.} 
As in Per-\texttt{FedAvg}, the first-order meta-gradient $v_i$ ideally requires a client-specific query dataset $\mathcal{D}_i^q$ to calculate an unbiased estimate $\nabla_{\phi} \tilde{f}_i (\phi_i;\mathcal{D}_i^q)$.
However, in \proposal, since the server does not have the required client data, we instead approximate $v_i$ by using the weight difference between a personalized model $\phi_i$ and global model  $\theta$ such that 
\begin{equation}
\label{eq:reptile}
    v_i = \nabla_{\phi} f_i (\phi_i)\approx \theta - \phi_i
\end{equation}
The intuition behind this method comes from the fact that the derivative of $\nabla_\phi \tilde{f}_i(\phi_i)$ in Eq.\eqref{eq:lossgradients} at a stationary point $\phi_i$ becomes sufficiently small.

A possible alternative to calculate $v_i$ at the server is to sample a query dataset $\mathcal{D}^q_s$ from server data distribution $p_s$ and calculate $\nabla_\phi \ell_i(\phi_i; \mathcal{D}^q_s)$. 
A potential drawback
% caveat 
in this approach is that $\mathcal{D}^q_s$ does not come from data distribution of client $i$. Our ablation study in Section~\ref{sec:eval} 
% will  
shows that the weight difference approximation is superior to direct calculation using server data.

%%%%%%%%%%%%%%%%%%%%%%%%%%%%%%%%%%%%%%%%%%%%%%%%%%%%%%%%%%%%
%%%%%%%%%%%%%%%%%%%%%%%%%%%%%%%%%%%%%%%%%%%%%%%%%%%%%%%%%%%%
% Second-order meta gradient
\textbf{Second-order meta-gradient.} 
To calculate $d_i$, instead of separately computing the Hessian $\nabla^2_{\phi} f_i(\phi_i)$, we approximate the entire Hessian-vector product $d_i$ by using Hessian-free estimation~\cite{convergencetheorymaml} as follows: 
\begin{eqnarray}
\label{eq:secondorderhessian}
    d_i &=& \nabla^2_\phi f_i(\phi_i) v_i  \nonumber \\
    &\approx& \frac{\nabla_\phi f_i(\phi_i+\delta v_i)-\nabla_\phi f_i(\phi_i-\delta v_i)}{2\delta}
\end{eqnarray}
This approximation produces an error of at most $\rho \delta \|v_i\|^2$, where $\rho$ is the parameter for Lipschitz continuity of the Hessian of $f$~\cite{convergencetheorymaml}. %TODO: Add which lemma?

Ideally, calculating unbiased estimates for the two first-order gradients $\nabla_\phi f_i(\phi_i+\delta v_i)$ and $\nabla_\phi f_i(\phi_i-\delta v_i)$ in Eq.(\ref{eq:secondorderhessian})  requires additional client-specific data. 
We take an alternative approach since the server does not have client data. The server samples $\mathcal{D}_s$ from its own data distribution $p_s$ and calculates  $\nabla_\phi \ell_i(\phi_i+\delta v_i; \mathcal{D}_s)$ and $\nabla_\phi \ell_i(\phi_i-\delta v_i; \mathcal{D}_s)$.

Note that we reuse the same dataset $\mathcal{D}_s$ to calculate $d_i$ for all clients in $S_t$. Given that the server data distribution $p_s$ is likely to be different from each client's data distribution, the quality of $d_i$ calculated using $\mathcal{D}_s$ may not be ideal. 
Nevertheless, we hypothesized that using the non-ideal second-order terms  would improve performance over disregarding the second-order terms altogether. The effectiveness of this approximation will be  empirically evaluated in Sections~\ref{sec:eval} and \ref{sec:stat}.

\textbf{Key differences.} Compared to Per-\texttt{FedAvg}~\cite{fallah2020personalized}, a recent meta-learning method for personalized ML, \proposal does not calculate meta-gradients, which is a computationally 
expensive operation, at resource-constrained clients but at the server. Moreover, \proposal is not restricted to one-step gradient update when calculating $\phi_i$ at clients but allows multi-step updates.
\texttt{pFedMe}~\cite{moreau_envelopes} and \texttt{FedProx}~\cite{li2020fedopt} are similar to \proposal in that the client-side problem in Eq.\eqref{eq: perfl_opt_problem_client} %use \eqref for equation references instead of (\ref{..})
includes a regularization term and each client utilizes multi-step gradient decent to obtain its optimized model $\phi_i$. On the other hand, \proposal enables meta-learning without more computation on clients.
Most importantly, \proposal actively utilizes the server to both aggregate personalized models and calculate computationally heavy meta-gradients by utilizing server data.
 
%%%%%%%%%%%%%%%%%%%%%%%%%%%%%%%%%%%%%%%%%%%%%%%%%%%%%%%%%%%%%%%%%%%%%%%%%%%%
\begin{table*}[htbp]
\centering
\vspace{-2ex}
\caption{Non-i.i.d. datasets and model architectures for federated learning benchmark~\cite{chaoyanghe2020fedml}.}
\label{table:datasets}
\resizebox{\textwidth}{!}{%
\begin{tabular}{l|ccccc}
\hline
\multirow{2}{*}{\textbf{Datasets}} & \textbf{\# of training} & \textbf{Non-i.i.d.} & \textbf{\# of partitions} & \textbf{\# of partitions} & \textbf{Baseline} \\
 & \textbf{samples} & \textbf{partition method} & \textbf{for clients} & \textbf{reserved server data} & \textbf{model architecture} \\ \hline\hline
Federated EMNIST & 671585 & realistic & 3230 & 0$-$170 & CNN (2 Conv + 2 FC) \\
CIFAR-100        & 50000  & Pachinko & 475  & 0$-$25 & ResNet-18 + group normalization \\
Shakespeare      & 16068  & realistic & 680  & 0$-$35 & RNN (2 LSTM + 1 FC) \\
StackOverflow    & 135818730 & realistic & 325354 & 0$-$17123 & RNN (1 LSTM + 2 FC) \\\hline
\end{tabular}}
\vspace{-2ex}
\end{table*}
%%%%%%%%%%%%%%%%%%%%%%%%%%%%%%%%%%%%%%%%%%%%%%%%%%%%%%%%%%%%%%%%%%%%%%%%%%%%
%%%%%%%%%%%%%%%%%%%%%%%%%%%%%%%%%%%%%%%%%%%%%%%%%%%%%%%%%%%%%%%%%%%%%%%%%%%%
\begin{figure*}[htbp]
    \centering
    \includegraphics[width=.93\linewidth]{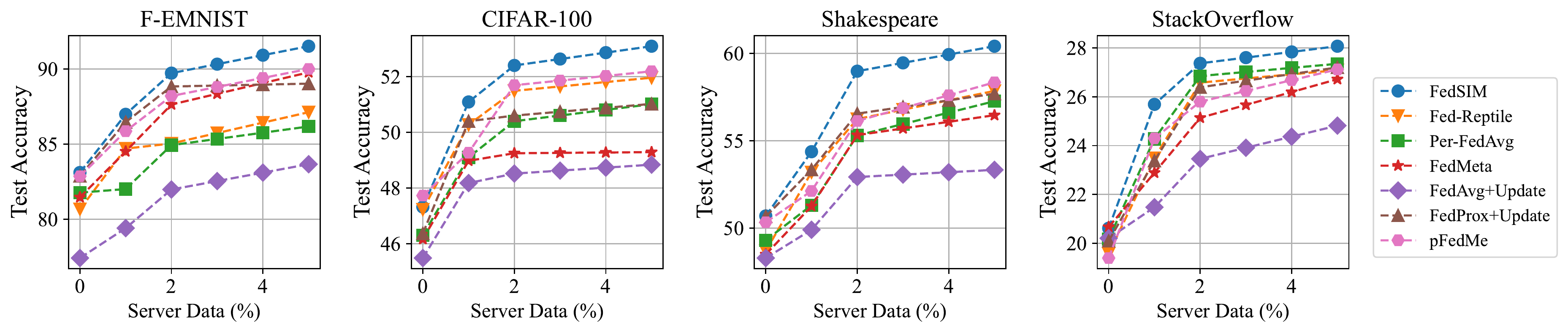}
    \vspace{-2ex}
    \caption{Effect of the proportion of server data on method performance when $E=5$.}
    \vspace{-3ex}
    \label{fig:acc_server}
\end{figure*}
%%%%%%%%%%%%%%%%%%%%%%%%%%%%%%%%%%%%%%%%%%%%%%%%%%%%%%%%%%%%%%%%%%%%%%%%%%%%

%%%%%%%%%%%%%%%%%%%%%%%%%%%%%%%%%%%%%%%%%%%%%%%%%%%%%%%%%%%%
%%%%%%%%%%%%%%%%%%%%%%%%%%%%%%%%%%%%%%%%%%%%%%%%%%%%%%%%%%%%
\subsection{Key Components for \texttt{Fed-SIM}}
%
% Overall, 
The key components
for the \proposal framework can be summarized as follows:
\vspace{-2ex}
\begin{itemize}[leftmargin=*]
    \item \textbf{Custom loss for local optimization}: Each client adds an $L_2$ regularization term to its loss function as in Eq.\eqref{eq: perfl_opt_problem_client} when optimizing a global model locally. This decouples meta gradient calculation (at the server) from local optimization history (at the clients).
    \item \textbf{First-order meta gradient calculation using weight differences}: Despite the existence of server data, the server calculates the first-order gradient $v_i$ using (client-specific) weight differences as in Eq.\eqref{eq:reptile} instead of using the server data. %With our custom loss, the former is not only easier to calculate, but also mathematically more accurate than the latter.
    \item \textbf{Second-order meta gradient calculation using server data}: The server calculates second-order gradient $d_i$ in a Hessian-Free way as in Eq.\eqref{eq:secondorderhessian}. The approximation requires the two terms calculated using server data $\mathcal{D}_s \sim p_s$ as $\nabla_\phi f_i(\phi_i+\delta v_i; \mathcal{D}_s)$ and $\nabla_\phi f_i(\phi_i-\delta v_i; \mathcal{D}_s)$. 
\end{itemize}
\vspace{-2ex}

With these components, 
\proposal ensures that data remains on the client while also ensuring that the calculation and communication done on the client is no more intensive than that done during standard federated learning. 
%At each round, each client focuses on optimizing its local model $\phi_i$ based on its private data. The server takes $\phi_i$ and uses its own dataset and computational resources to calculate meta gradients $\nabla _\theta F_i(\theta_{t-1})$, which is used to update the adaptable global model $\theta_{t-1}$.
\vspace{-1ex}

\section{Experiments}
\label{sec:eval}

The goal of our experiments is to evaluate (\romannumeral 1) the performance of \proposal compared with existing methods on personalized FL with non-i.i.d. client data, (\romannumeral 2) the convergence and computational overhead of \proposal, and (\romannumeral 3) the effectiveness of the three key components for \proposal. All our experiments were simulated using a server comprising four NVIDIA RTX 3900 GPUs and two Intel Xeon Silver CPUs. To our knowledge, this section also serves as the most comprehensive empirical study on personalized FL.

%%%%%%%%%%%%%%%%%%%%%%%%%%%%%%%%%%%%%%%%%%%%%
%%%%%%%%%%%%%%%%%%%%%%%%%%%%%%%%%%%%%%%%%%%%%
\subsection{Experimental Design}
\label{sec:experiment_design}

\textbf{Benchmarks.} 
We compare \proposal with other personalized FL methods based on \textit{optimization-based meta-learning}, \texttt{FedMeta}~\cite{chen2019federated}, \texttt{Fed-Reptile}~\cite{jiang2019improving}, \texttt{Per-FedAvg} (FO)\footnote{\texttt{Per-FedAvg (FO)} is the first-order approximation of \texttt{Per-FedAvg}, which makes clients compute first-order meta gradients but disregard second-order terms. The full version of \texttt{Per-FedAvg} is the same as \texttt{FedMeta}.}~\cite{fallah2020personalized}, and \texttt{pFedMe}~\cite{moreau_envelopes}, and also regular FL methods \texttt{FedAvg}~\cite{mcmahan2017communicationefficient} and \texttt{FedProx}~\cite{li2020fedopt}. 
Note that since FL is a newly growing research field, existing work have used their own benchmarks to evaluate their respective methodologies, with their own methods of splitting data in a non-i.i.d. manner, which made it difficult to provide a fair comparison in performance.

To mitigate this problem, the authors of FedML~\cite{chaoyanghe2020fedml} opened a research library including benchmarks for federated learning. Thus, we use four non-i.i.d. datasets, Federated EMNIST~\cite{caldas2019leaf}, CIFAR-100~\cite{cifar10}, Shakespeare~\cite{mcmahan2017communicationefficient}, and StackOverflow~\cite{StackOverflow}, and train a standardized neural network for each dataset with experiments constructed as suggested in~\cite{chaoyanghe2020fedml}. 
While three of these datasets are naturally partitioned with a non-i.i.d. distribution, CIFAR-100 is partitioned using Pachinko Allocation Method as in~\cite{reddi2020adaptive}. The exact specifications are summarized in Table \ref{table:datasets}.

%%%%%%%%%%%%%%%%%%%%%%%%%%%%%%%%%%%%%%%%%%%%%%%%%%%%%%%%%%%%%%%%%%%%%%%%%%%%
\begin{figure*}[htbp]
    \centering
    \includegraphics[width=.99\linewidth]{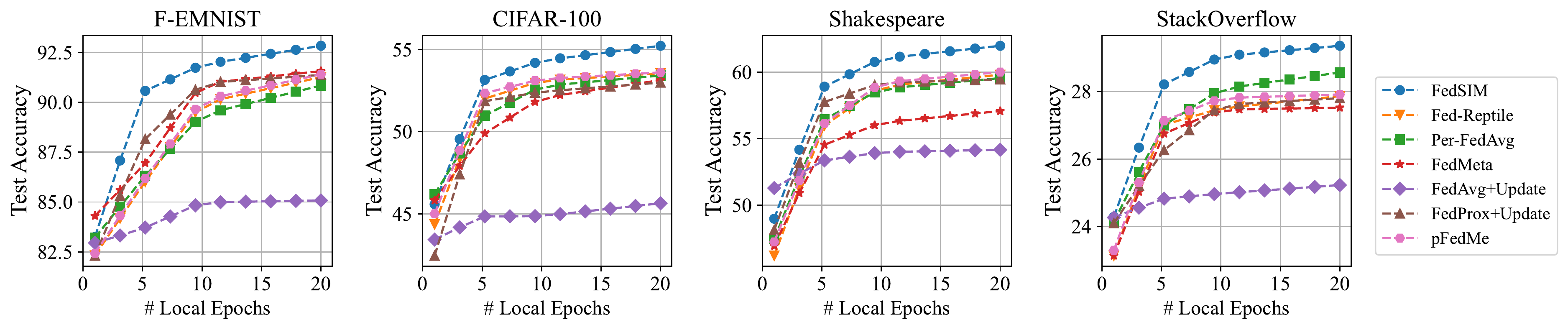}
    \vspace{-2ex}
    \caption{Effect of the number of local epochs $E$ for fine-tuning $\phi_i$ when server data proportion=5\%.}
    \vspace{-2.5ex}
    \label{fig:acc_epochs}
\end{figure*}

%%%%%%%%%%%%%%%%%%%%%%%%%%%%%%%%%%%%%%%%%%%%%
%%%%%%%%%%%%%%%%%%%%%%%%%%%%%%%%%%%%%%%%%%%%%
\textbf{Server Data Simulation.} 
For each dataset, we randomly sample 5\% of the non-i.i.d. data partitions and reserve them as server data, while using the remaining 95\% partitions as client datasets. It is important to note that while all the methods use server data for training an initial model, only \proposal uses server data during the actual FL process. Given that \proposal takes advantage of server data, we also experimented with different amounts of server data.  %when training the other methods to keep our experiments as fair as possible. %In the case of the other methods, server data is used only for training an initial model but not during the federated learning process.

Furthermore, when running \texttt{Per-FedAvg} (FO) and \texttt{FedMeta}, 80\% and 20\% of each client's training data are allocated as the client's support and query datasets respectively for local calculation of meta gradients.

%%%%%%%%%%%%%%%%%%%%%%%%%%%%%%%%%%%%%%%%%%%%%
%%%%%%%%%%%%%%%%%%%%%%%%%%%%%%%%%%%%%%%%%%%%%
\textbf{Training Process.}
Training is carried out with $m=10$ as in~\cite{li2020fedopt, reddi2020adaptive}, such that $m$ clients are randomly sampled in each round to perform local optimization.
Test accuracy is evaluated every round by sampling $m$ clients, deploying the current global model, fine-tuning (personalization) on each client's training data, and finally averaging the validation accuracy of all clients.
Note that since \texttt{FedProx} and \texttt{FedAvg} do not provide a personalization step, we add an \textit{update} step to simulate personalization of the global model.

%%%%%%%%%%%%%%%%%%%%%%%%%%%%%%%%%%%%%%%%%%%%%%%%%%%%%%%%%%%%%%%%%%%%%%%%%%%%
\begin{table}[t]
\centering
\vspace{-1ex}
\caption{Best performance of each method on each dataset.}
\label{table:perfed_acc_server}
\resizebox{\linewidth}{!}{%
\begin{tabular}{l|cccc}
\hline
\textbf{Methodologies} & \textbf{Fed. EMNIST} & \textbf{CIFAR-100} & \textbf{Shakespeare} & \textbf{StackOverflow} \\\hline\hline

\texttt{FedProx} + update & 88.03& 51.82& 57.55& 27.41\\
FedAvg + update & 83.66 & 41.49 & 54.28 & 25.21 \\
\hline
\texttt{pFedMe} & 89.26& 52.31& 59.02& 27.91\\
\texttt{Per-FedAvg} (FO) & 86.17 & 50.99 & 58.34 & 27.95 \\
\texttt{FedMeta} & 89.77 & 46.29 & 51.46 & 26.72 \\
\texttt{Fed-Reptile} & 85.84 & 51.96 & 55.85 & 27.29 \\
\textbf{\proposal (ours)} & \textbf{91.51 $\pm$ 0.273} & \textbf{53.09 $\pm$ 0.104} & \textbf{60.81 $\pm$ 0.329} & \textbf{28.17 $\pm$ 0.171} \\\hline
\end{tabular}}
\vspace{-2ex}
\end{table}

\begin{figure*}[t]
    \centering
    \includegraphics[width=.9\linewidth]{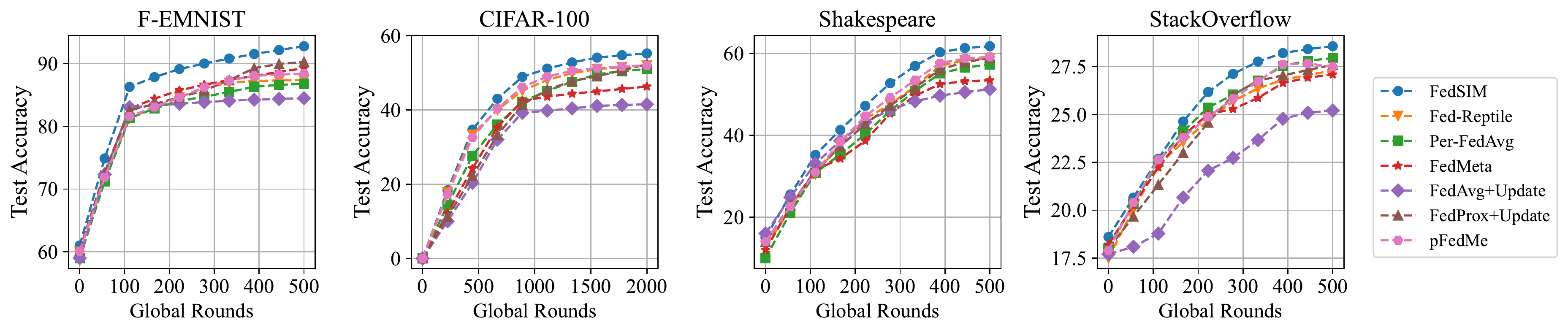}
    \vspace{-2ex}
    \caption{Test accuracy as communication round increases when $E=10$ and server data proportion is 5\%.}
    \vspace{-2ex}
    \label{fig:comm_histogram}
\end{figure*}
%%%%%%%%%%%%%%%%%%%%%%%%%%%%%%%%%%%%%%%%%%%%%%%%%%%%%%%%%%%%%%%%%%%%%%%%%%%%

%%%%%%%%%%%%%%%%%%%%%%%%%%%%%%%%%%%%%%%%%%%%%%%%%%%%%%%%%%%%%%%%%%%%%%%%%%%%
\subsection{Method Performance}
\label{sec:method_performance}

\textbf{Effects of Server Data Proportion.} 
Figure \ref{fig:acc_server} shows the average test accuracy of various personalized FL methods with varying amounts of server data. Note that while the amount of server data varies from 0 to 5\%, 95\% of the entire dataset are always allocated as clients. 
The results show that more server data results in better accuracy in all methods, implying that training an initial model using server data improves performance. \texttt{FedAvg} shows the worst performance since it does not train an adaptable (personalizable) model. Although the performance of other five conventional methods vary by data settings, \proposal always provides the best accuracy once server data is given.

In particular, with 5\% server data, our method's performance exceeds all other values in every dataset. As shown in Table~\ref{table:perfed_acc_server}, when comparing the best values in each dataset, \proposal provides 0.22$-$2.57\% higher accuracy than the next best methods. 
This verifies that \proposal's meta-gradient computation is an effective way for using server data \textit{during} the FL process even when server data is not representative of the entire dataset.

Note that server data is not ideal since conventional MAML requires task (client)-specific datasets for meta gradients. However, our results suggest that if the client datasets are not given to the server due to privacy concerns, calculation of second-order meta gradients using the server data can be a good alternative rather than giving up the second-order terms as in \texttt{Per-FedAvg} (FO), \texttt{pFedMe}, and \texttt{Fed-Reptile}. When there is no server data, \proposal cannot calculate Hessian estimates, essentially becoming the same as \texttt{pFedMe}. In this setting, however, \proposal still outperforms both \texttt{Fed-Reptile} and \texttt{FedProx}, showing that the implementation of both a custom loss and first-gradient estimates results in more accurate meta-gradients by preventing local model divergence.

Furthermore, \proposal shows that utilizing the server to directly calculate meta gradients is more effective than simply averaging locally trained meta models as in \texttt{FedMeta}. Note that \texttt{FedMeta} enables each client to calculate full meta gradients including second-order terms on its client-specific dataset when optimizing its local model, which requires heavy computation on clients but turns out to be not effective for improving the global model.

%%%%%%%%%%%%%%%%%%%%%%%%%%%%%%%%%%%%%%%
%%%%%%%%%%%%%%%%%%%%%%%%%%%%%%%%%%%%%%%
\textbf{Effects of Local Epochs.} 
Figure \ref{fig:acc_epochs} shows test accuracy of the same methods with 5\% server data and varying local epochs $E$.
While all the methods show better accuracy as $E$ increases, \proposal experiences remarkable improvement when $E$ increases from 1 to 5 and regularly outperforms all the other methods when $E \ge 5$. In each dataset, the best accuracy value is given by \proposal with $E=20$, showing an 1.09$-$2.57\% increase in accuracy compared to the second highest values.
%
%It is important to note that \proposal achieves such performance improvement because the custom loss function in \proposal decouples meta gradient calculation at the server from each client's local optimization history. 

%%%%%%%%%%%%%%%%%%%%%%%%%%%%%%%%%%%%%%
%%%%%%%%%%%%%%%%%%%%%%%%%%%%%%%%%%%%%%
\subsection{Resource Efficiency}

\begin{figure*}[t]
    \centering
    \includegraphics[width=.9\linewidth]{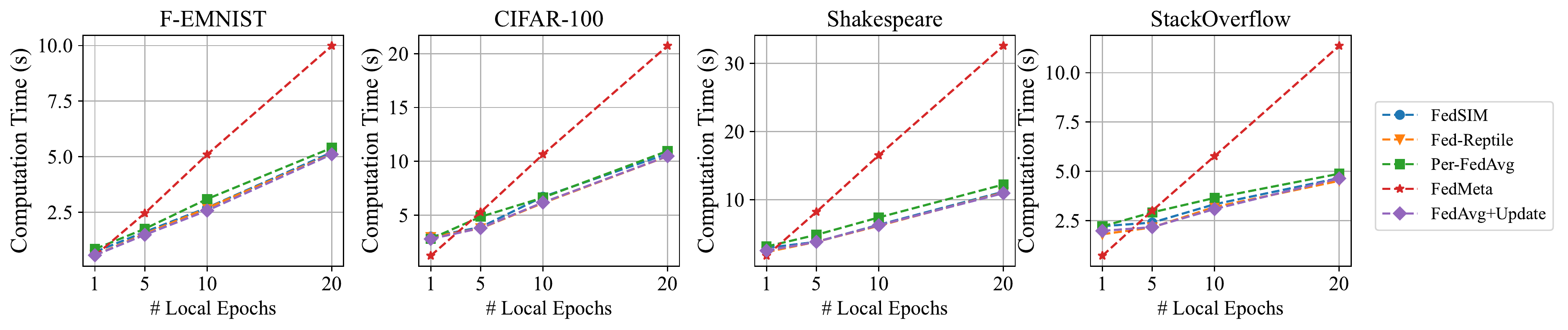}
    \vspace{-2ex}
    \caption{Effect of epochs on average per-client, per-round computation time (seconds).}
    \vspace{-2ex}
  \label{fig:comp_histogram}
\end{figure*}

Next, we evaluate the resource efficiency of \proposal in terms of local computation and communication overhead.
Figure \ref{fig:comm_histogram} plots test accuracy as communication round ($t$) increases. While \proposal achieves the highest accuracy in all cases, it achieves the next best accuracy in 34.2\%, 11.38\%, 19.44\%, 20.07\% fewer communication rounds for each respective dataset due to the use of more accurate meta gradients for model updates. Given that all the methods require the same communication overhead in each round (i.e., dissemination of $\theta$ and aggregation of $\phi_i$), fewer rounds entail less communication overhead.

Figure \ref{fig:comp_histogram} 
shows the
average client computation time for local optimization in each round. The computation time of \texttt{FedMeta} (or \texttt{Per-FedAvg}) quickly increases with local epochs due to local calculation of second-order meta gradients. \texttt{Per-FedAvg} (FO) ignores the second-order terms but still calculates first-order meta gradients locally, resulting in the second-longest computation time. On the other hand, \proposal shows a modest increase in client computation time as local epoch increases, similar to \texttt{FedAvg} that does not provide personalization, since meta gradients are calculated at the server. Overall, \proposal not only trains a more accurate model but also does so resource-efficiently.

%%%%%%%%%%%%%%%%%%%%%%%%%%%%%%%%%%%%%%%%%%
%%%%%%%%%%%%%%%%%%%%%%%%%%%%%%%%%%%%%%%%%%
\subsection{Ablation Studies}

\begin{table*}[hbt]
\vspace{-2ex}
\centering
\caption{\% accuracy of \proposal variants with $E=5$ and 5\% server data.}
\label{table:ablation}
\resizebox{.85\textwidth}{!}{%
\begin{tabular}{c|ccc||cccc}
\hline
 \multirow{2}{*}{\textbf{Methodologies}} & \textbf{Loss} & \textbf{FO meta} & \textbf{SO meta} & \textbf{Federated} &  \multirow{2}{*}{\textbf{CIFAR-100}} &  \multirow{2}{*}{\textbf{Shakespeare}} & \multirow{2}{*}{\textbf{StackOverflow}} \\
 & \textbf{function} &  \textbf{gradient}s & \textbf{gradients} & \textbf{EMNIST} &  &  &  \\ \hline\hline
\textbf{\proposal-var1} & basic & weight diff & server & 90.80 & 49.73 & 55.68 & 26.81 \\
\textbf{\proposal-var2} & custom & server data & server & 77.78 & 48.45 & 53.49 & 25.76 \\
\textbf{\proposal-var3} & custom & weight diff & x & 85.86 & 51.01 & 57.26 & 27.12 \\
\textbf{\proposal} & custom & weight diff & server & 91.51 & 53.09 & 60.81 & 28.17 \\\hline
\end{tabular}}
\vspace{-4ex}
\end{table*}

% Given that server data distribution may have dissimilar data distributions from each client,
Given that the distribution of server data is dissimilar to that of each client's data,
using server data without caution may end up with performance degradation. To this end, we evaluate if each key component of \proposal actually contributes to its performance, namely the (\romannumeral 1) loss function, (\romannumeral 2) first-order (FO) meta gradient calculation, and (\romannumeral 3) second-order (SO) meta gradient calculation. 
We made three variants of \proposal, \proposal-var1 that uses basic loss function without $L_2$ regularization, \proposal-var2 that calculates FO meta gradients using server data instead of weight difference (i.e., $\nabla_\phi \tilde{f}_i(\phi_i; \mathcal{D}^q_s)$ where $\mathcal{D}^q_s \sim p_s$), and \proposal-var3 that disregards SO meta gradients.

Table~\ref{table:ablation} shows the performance of these variants. Comparison with \proposal-var3 verifies that although calculating SO meta gradients using client-independent server data is not theoretically ideal, using the SO terms still results in significantly better performance than relying only on FO meta gradients. The \proposal-var2 case shows, however, that the non-ideal server data causes severe performance degradation when used for FO meta gradient calculation; using (client-specific) weight differences is a better choice in case of calculating FO meta gradients. In addition, \proposal-var1 proves that using a custom loss to decouple local optimization history from meta gradients and calculating meta gradients based on $\phi_i$ (locally optimized model) rather than $\theta$ (previous meta model) result in more useful meta gradients. 
Overall, the results verify that each of the key components of \proposal highly impacts model accuracy.

\section{Data Dissimilarity Analysis} %Statistical vs. Data Dissimilarity?
\label{sec:stat}

We can see in practical FL scenarios that although clients may have non-i.i.d. data, the data distributions of clients are not entirely unrelated. Thus,
%which is the reason why the global model converges using FL algorithms. 
prior work such as \texttt{FedProx}~\cite{li2020fedopt}, \texttt{Per-FedAvg}~\cite{fallah2020personalized}, and \texttt{pFedMe}~\cite{moreau_envelopes}, perform convergence analyses on the global model by assuming that both data distributions and local gradients have bounded dissimilarity among clients. 
\proposal makes a similar assumption
%is built on a similar assumption that server data distribution has bounded dissimilarity with client data distributions 
that both server data distributions and meta gradients calculated using server data have bounded dissimilarity. Thus, in this section, we conduct 
experiments to evaluate the effect of distributional deviation
% statistical analyses 
between server and client data.

First, we investigate data dissimilarity between the server and clients in non-i.i.d. data settings with varying amount of server data. Next, we empirically observe that the variance of data dissimilarity between the clients and the server is an appropriate measure of model performance in \proposal. %\MHO{(This last sentence is a little vague.)}

%statistical arguments discussing how the model performs based on the relationship between the distribution of server data and the distribution of client data.

\begin{figure}[t]
    \includegraphics[width=\linewidth]{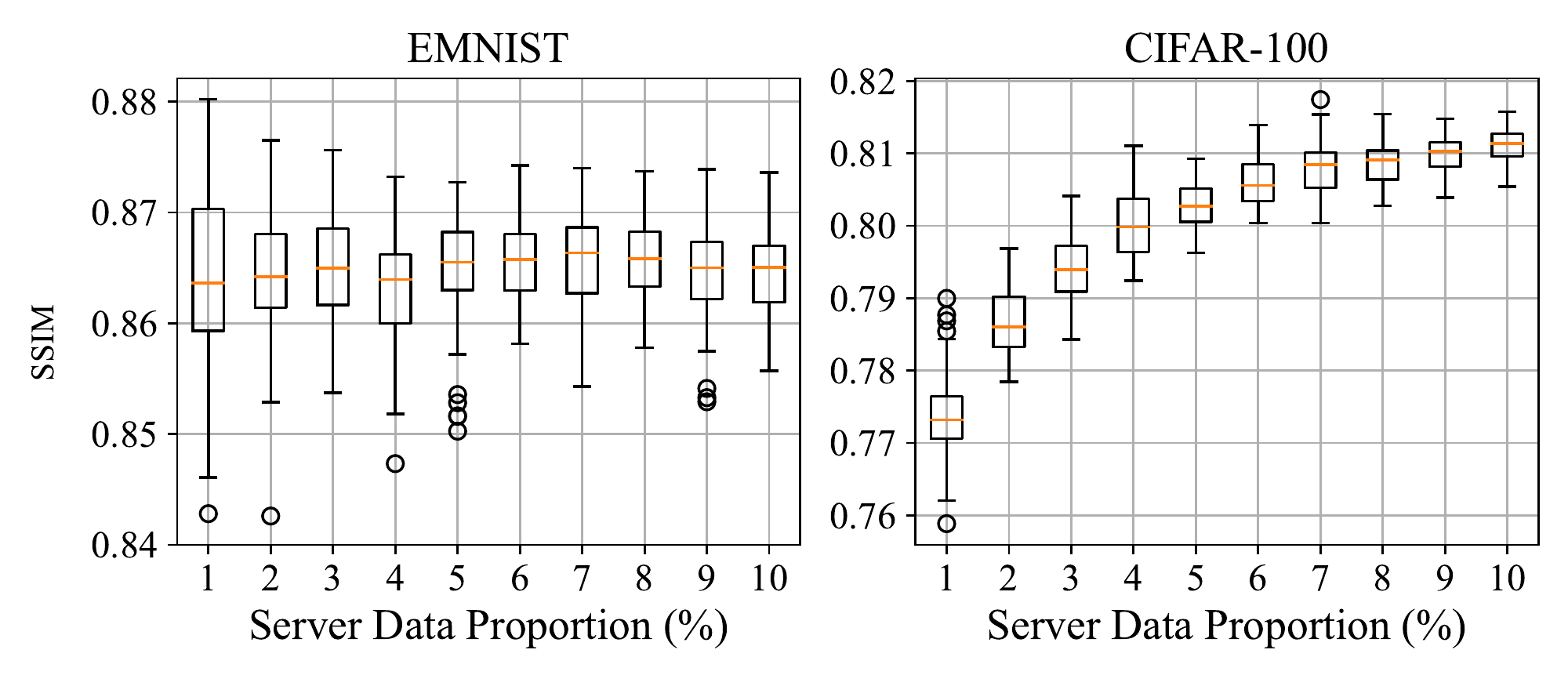}
    \vspace{-5ex}
    \caption{Effects of varying amount of server data on image similarity distributions.}
    \vspace{-3ex}
    \label{fig:ssim_wd}
\end{figure}

%%%%%%%%%%%%%%%%%%%%%%%%%%%%%%%%%%%%%%%%%%%%%%
%%%%%%%%%%%%%%%%%%%%%%%%%%%%%%%%%%%%%%%%%%%%%%
\textbf{Distribution Comparison.} 
To investigate data dissimilarity, we randomly sample a small percentage of data from two image datasets, Federated-EMNIST and CIFAR-100, as in Section~\ref{sec:experiment_design} to simulate server data.
The average image distribution of the server data is then compared to each of the remaining clients, using a Structural Similarity Index (SSIM)~\cite{ssim} to compare the image data. % and Wasserstein Distance (WD) to compare the label distributions. 
Thus, each comparison produces $SSIM_{(i,j)}$ %a tuple of $(SSIM_{(i,j)}, WD_{(i,j)})$ 
for $i \in \mathcal{D}_s$ and $j \in \mathcal{D}$ where $\mathcal{D}$ is the set of all clients in a dataset and $\mathcal{D}_s$ is the set of server data. This process is repeated many times for each proportion of server data, resulting in boxplots in Figure \ref{fig:ssim_wd}.

%\begin{figure}[t]
%    \includegraphics[width=\linewidth]{Graphs/wasserstein.png}
%    \vspace{-4.5ex}
%    \caption{Effects of varying amount of server data on label-wise Wasserstein distance.}
%    \vspace{-2.5ex}
%    \label{fig:wasserstein}
%\end{figure}

Figure \ref{fig:ssim_wd} shows different trends of data similarity in the two datasets. Regarding CIFAR-100, there is a noticeable increase in SSIM with more server data. This is contrary to SSIM in EMNIST, which remain fairly consistent.
We hypothesize that this is due to the fact that EMNIST is a relatively simple dataset, not only represented in grayscale but also consisting of handwritten letters that hardly differ by client, which leads to fast saturation of data similarity with only a small amount of server data. On the other hand, CIFAR-100 provides far more diverse images which require more server data such that data similarity can converge (albeit at a lower SSIM than EMNIST), which is more representative of real-world images.

Despite the differences, in both datsets, the general trend of the variance of the similarity metrics decrease with more server data. In addition, SSIM is higher than 0.75 even when server data proportion is 1\%, showing that server and client data distributions are not entirely unrelated.

\begin{figure}[t]
\centering
    \includegraphics[width=\linewidth]{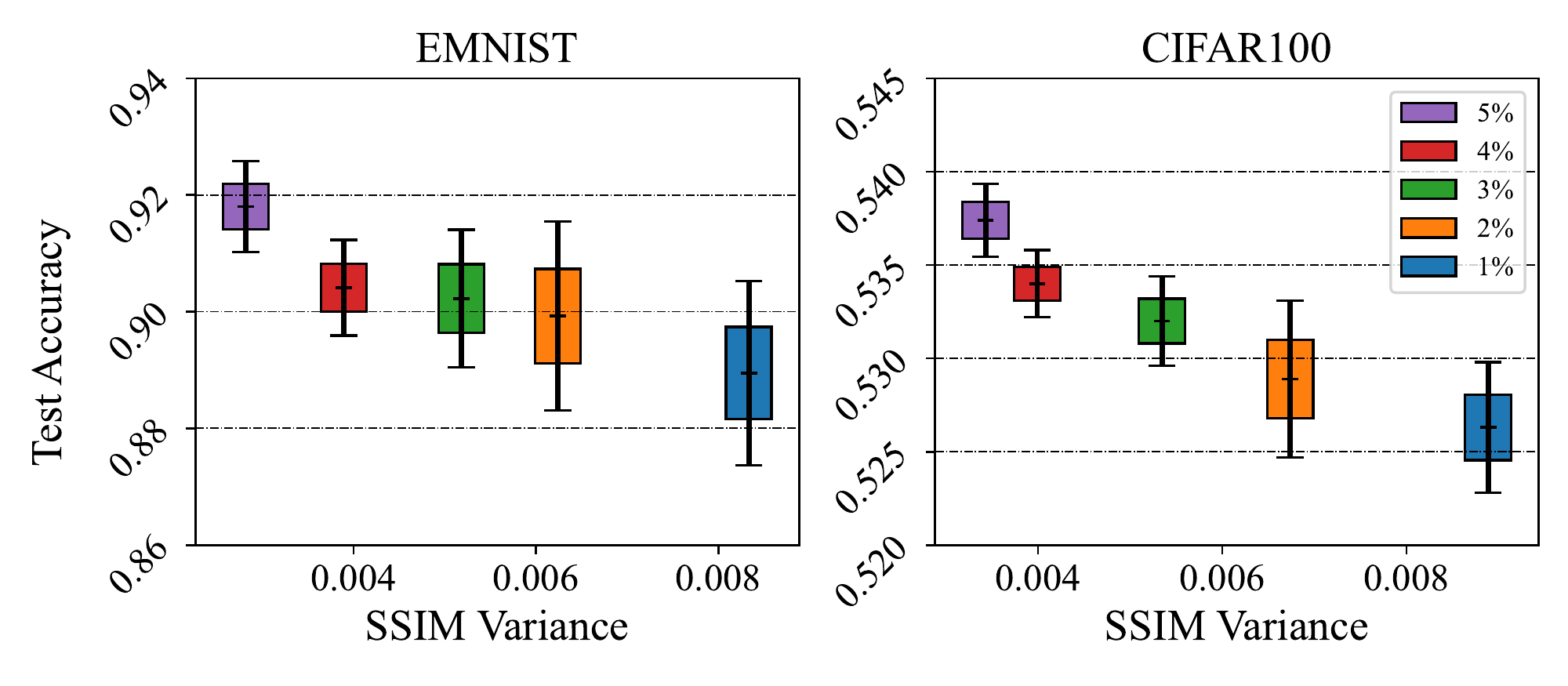}
    \vspace{-5ex}
    \caption{Change in model performance in relation to the change in structural similarity of images.}
    \vspace{-3ex}
    \label{fig:model_performance_ssim}
\end{figure}

%%%%%%%%%%%%%%%%%%%%%%%%%%%%%%%%%%%%%%%%%%%%%%
%%%%%%%%%%%%%%%%%%%%%%%%%%%%%%%%%%%%%%%%%%%%%%
\textbf{Relationship with Performance.}
We then analyze the impact of average image similarity on model performance, as seen in Figure \ref{fig:model_performance_ssim}. This figure shows the relationship between \proposal performance on the EMNIST and CIFAR-100 datasets and the variance in image similarity (SSIM) between the server data and each of the client data. Data points were collected by training the models various times with different amounts of server data. Then, the server data used during each learning process is used to calculate image similarity with the remaining client data. Finally, we measure the variance of image similarity for each data point, and plot it with respect to the distribution of model accuracy. 

Here, the middle of each box represents the mean variance in image similarity for each server data proportion, while the error bars represent the variance in test accuracy. The graphs show that there is a negative correlation between SSIM variance and model performance. More server data results in better accuracy due to its correlation with SSIM variance. This implies that even with the same amount of server data, model performance can depend on the method by which the server dataset is constructed.
\vspace{-1ex}
\section{Discussion}

In this paper, we investigate a practical problem setting of FL, personalized federated learning with server data. We adapt the meta-learning process to create \proposal where meta-gradients are calculated using the server to improve model performance and reduce client computational overhead. We show that \proposal solves the proposed FL problem by first performing local optimization using a \textit{custom loss function} with a regularization term, and then using \textit{server data} with these locally optimized models to calculate the required gradients. We also provide a variety of numerical experiments and ablations to illustrate the performances of our method compared with existing methods in personalized FL. Finally, we present empirical analyses on the distribution of server data and its impact on performance.

While we focus on personalized FL and meta learning, we believe that this work opens up an interesting avenue in the FL regime that investigates how powerful server and its data can contribute to federated learning process effectively.

\bibliography{background.bib}
\bibliographystyle{icml2022}

\newpage
\appendix
\onecolumn

\section{Qualitative Comparison with Other Methods}
\label{appendix:comparison}

Table~\ref{table:compare_perfed} compares various aspects of existing methods in personalized federated learning% and meta learning, respectively, 
including \proposal. 
Compared to existing methods in personalized federated learning, \proposal is differentiated in that it (1) utilizes a custom loss for local optimization and (2) calculates full meta-gradient at the server without additional communication overhead: first-order meta gradient using weight difference and second-order gradient using Hessian-free approximation and proxy data.
%
%Compared to existing methods in meta learning, \proposal's uniqueness is that it (1) performs task-specific optimization at clients without sharing their raw data with the server and (2) does not use task-specific query datasets for calculating meta gradients but utilizes model weights for first-order meta gradients and task-independent proxy data for second-order meta gradients.

\begin{table}[h]
\centering
\caption{Qualitative Comparison of existing methods in personalized federated learning}
\label{table:compare_perfed}
\resizebox{\textwidth}{!}{%
\begin{tabular}{c|c|ccccc}
\hline
\multicolumn{2}{c|}{\multirow{2}{*}{\textbf{Methodologies}}} & \textbf{FedAvg} & \textbf{Per-FedAvg} & \textbf{FedMeta} & \textbf{Fed-Reptile} & \textbf{\proposal} \\
\multicolumn{2}{c|}{}  & \cite{mcmahan2017communicationefficient}  & \textbf{(FO)}~\cite{fallah2020personalized} & \cite{chen2019federated} & \cite{jiang2019improving}  & (ours) \\\hline\hline
\textbf{Local}           & Loss function    & general & general & general & general & L2 regularization \\\cline{2-7}
\textbf{(task-specific)} & Where to & \multicolumn{5}{c}{\multirow{2}{*}{client}} \\
\textbf{optimization}    & compute  &         &         &         &         &              \\\hline\hline
                       & Required    & \multirow{2}{*}{N/A}  & entire history, & entire history, & \multirow{2}{*}{final weights} & \multirow{2}{*}{final weights} \\
                       & information &                       & query datset    & query dataset   &                                &  \\\cline{2-7}
\textbf{First-order}   & \multirow{3}{*}{Method} & \multirow{3}{*}{N/A} & exact       & exact       & weight        & weight \\
\textbf{meta gradient} &                         &                      & calculation & calculation & difference    & difference \\
\textbf{(outer loop)}  &                         &                      &             &             & approximation & approximation     \\\cline{2-7}
                       & Where to  & \multirow{2}{*}{N/A} & \multirow{2}{*}{client} & \multirow{2}{*}{client} & \multirow{2}{*}{server} & \multirow{2}{*}{server} \\
                       & compute &  &  &  &  &  \\\hline\hline
                       & Required    & \multirow{2}{*}{N/A} & \multirow{2}{*}{N/A} & entire history, & \multirow{2}{*}{N/A} & final weights, \\
\textbf{Second-order}  & information &                      &                      & query dataset   &                      & proxy data \\\cline{2-7}
\textbf{meta gradient} & \multirow{2}{*}{Method} & \multirow{2}{*}{N/A} & \multirow{2}{*}{N/A} & exact       & \multirow{2}{*}{N/A} & Hessian-free   \\
\textbf{(outer loop)}  &                         &                      &                      & calculation &                      & approximation  \\\cline{2-7}
                       & Where to & \multirow{2}{*}{N/A} & \multirow{2}{*}{N/A} & \multirow{2}{*}{client} & \multirow{2}{*}{N/A} & \multirow{2}{*}{server} \\
                       & compute  &     &     &        &     &        \\\hline\hline
\multicolumn{2}{c|}{\textbf{Where training data is stored}} & \multicolumn{5}{c}{Mostly on clients, (optionally) small amount of proxy data on server} \\\hline
\end{tabular}}
\end{table}

\section{Experiment Details}
\label{appendix:experiment}

In this section, we provide additional details of the experimental set-up for the experiments in Section \ref{sec:eval}. We used federated versions of vision datasets EMNIST~\cite{emnist} and CIFAR-100~\cite{cifar10}, alongside language modeling datasets Shakespeare~\cite{mcmahan2017communicationefficient} and StackOverflow~\cite{StackOverflow}.

We train our model such that in each communication round, 10 clients are sampled, the model is trained using each respective methodology on each client's \textit{training dataset}. At the end of each communication round, we sample 10 different individuals, where each client is first fine-tuned using standard training with no custom loss, and tests on its testing dataset using its own fine-tuned model. We take the average of the client's test accuracy to evaluate the model's performance. Note that we use $\lambda = 1$ and $\delta = 0.25$.

Thus, each communication round can be summarized as the following. 
\begin{enumerate}
    \item Sampling phase: where a number of clients are chosen from the entire client pool, each with their own unique data randomly sampled from the training dataset
    \item Training phase: where the model is trained to quickly adapt to each unique client
    \item Testing phase: where the model is tested on a new client with data from the test dataset

\end{enumerate}

Furthermore, a summary of the hyperparameters we used for each dataset is given in Table~\ref{table:hyperparameters}. Note that we fix the batch size at a per-task level given the large number of hyperparameters to tune and to avoid conflating variables.

\begin{table*}[ht]
\label{table:hyperparameters}
\caption{Summary of hyperparameters used for each task}
\centering
\begin{tabular}{lllll} 
\toprule
Hyperparameters      & Federated EMNIST & CIFAR-100    & Shakespeare  & StackOverflow  \\ 
\hline
Client Optimizer     & Adam             & Adam         & Adam         & Adam           \\
Client Scheduler     & x                & x            & x            & x              \\
Client Learning Rate & 0.01             & 0.001        & 0.001        & 0.001          \\
Server Optimizer     & SGD              & SGD          & SGD          & SGD            \\
Server Scheduler     & Linear Decay     & Linear Decay & Linear Decay & Linear Decay   \\
Server Learning Rate & 0.25             & 0.25         & 0.25         & 0.25           \\
Batch Size           & 20               & 20           & 4            & 16             \\
\hline
\end{tabular}
\end{table*}

\subsection{EMNIST}

EMNIST~\cite{emnist} consists of images of digits and upper and lower case English characters, with 62 total classes. The federated version of EMNIST~\cite{caldas2019leaf} partitions the digits by their author. The dataset has natural heterogeneity stemming from the writing style of each person. We perform an character recognition task using this dataset, with a full description of the model in Table \ref{EMNIST model table}.

Federated EMNIST is partitioned in a manner such that 3,400 individuals constitute a separate client, with each client having an individual training dataset and a testing dataset. Thus, a testing round for EMNIST consists of sampling a user, training on the user's handwriting style, and testing on the individual testing dataset for that particular user. 

\begin{table*}[ht]
\centering
\caption{Federated EMNIST model architecture.}
\label{EMNIST model table}
\begin{tabular}{lllll}
\toprule
Layer     & Output Shape & \# of Trainable Parameters & Activation & Hyperparameters               \\
\hline
Input     & (28,28,1)    & 0                          &            &                               \\
Conv2d    & (26,26,32)   & 320                        &            & kernel size=3; strides=(1,1)  \\
Conv2d    & (24,24,64)   & 18496                      & ReLU       & kernel size=3; strides=(1,1)  \\
MaxPool2d & (12,12,64)   & 0                          &            &                               \\
Dropout   & (12,12,64)   & 0                          &            & $p = 0.25$                      \\
Flatten   & 9216         & 0                          &            &                               \\
Dense     & 128          & 1179776                    &            &                               \\
Dropout   & 128          & 0                          &            & $p = 0.5$                       \\
Dense     & 62           & 7998                       & softmax    &                               \\
\hline
\end{tabular}
\end{table*}

\subsection{CIFAR-100}

CIFAR-100 consists of images with RGB channels of 32x32 pixels each. Each pixel is represented by an unsigned int8. As is standard with CIFAR datasets, we perform preprocessing on the training images. For training images, we augment the data by performing a random horizontal flip. We then scale the pixel values such that each pixel value lies between $[0,1]$. We train a modified ResNet-18 model, where the batch normalization layers are replaced by group normalization layers. 

Our model trains on a federated version of CIFAR-100 as proposed in~\cite{reddi2020adaptive}, where the authors apply a two step latent Dirichlet allocation (LDA) process by first randomly partitioning the data to reflect the  "coarse" and "fine" labels structure of CIFAR-100 by using the Pachinko Allocation Method (PAM), and finally creating a federated dataset using LDA with a parameter of 0.1. Using this method, the authors of~\cite{reddi2020adaptive} create a training dataset consisting of 500 clients and a testing dataset consisting of 100 clients.

Although we train our model using the 500 training clients, we needed to slightly modify the testing dataset in order to allow fine-tuning of the model when deployed. To do so, for each test client, we split the client dataset into a fine-tuning dataset and a validation dataset consisting of 80 and 20\% of the data respectively. By doing so, when testing, we sample ten clients from the test client space, optimize the models on the fine-tuning dataset for each client, and evaluate the models using each client's respective validation datasets.

\subsection{Shakespeare}

Shakespeare is a language modeling dataset built from the collective works of William Shakespeare and first used in~\cite{mcmahan2017communicationefficient} as a federated learning task. The dataset consists of 715 "actors", each with their own distinct method of talking. Each client's lines are partitioned into training and test sets. The natural language processing task here is to perform next character prediction. To do so, we use an RNN that takes a series of 80 characters as input, passes it through the model, and outputs a sequence of characters formed by shifting the input sequence by one. This way, the last character is the new character we are actually trying to predict.

The model architecture for the Shakespeare character prediction task is shown in Table \ref{table:shakespeare_architecture}.

\begin{table*}[h]
\caption{Shakespeare model architecture.}
\label{table:shakespeare_architecture}
\centering
\begin{tabular}{lll} 
\toprule
Layer     & Output Shape & \# of Trainable Parameters  \\ 
\hline
Input     & 80           & 0                           \\
Embedding & (80, 8)      & 720                         \\
LSTM      & (80, 256)    & 271360                      \\
LSTM      & (80, 256)    & 575312                      \\
Dense     & (80, 90)     & 23130                       \\
\hline
\end{tabular}
\end{table*}

\subsection{StackOverflow}

StackOverflow is a language modeling dataset consisting of questions and answers from the site Stack Overflow. The dataset contains 342,477 unique users which we use as clients. We perform next-word prediction on this dataset. 

\textbf{Preprocessing}
For this task, we restrict the dataset to 10,000 most frequently used words, restrict each client such that they have at most 1000 sentences in their dataset. We then truncate or pad the data such that each sentence has 21 words (20 words are used as the input sequence, and the last word is the predicted output). We then represent the sentence as a sequence of indices corresponding to the 10,000 most frequently used words.
Our RNN model embeds these sequences into a learned 96-dimensional space. It then feeds the embedded words into a single LSTM layer, followed by two densely connected layers with a softmax activation at the end. The model architecture for the StackOverflow next word prediction task is shown in Table \ref{table:stack_architecture}. 

\begin{table*}[h]
\caption{StackOverflow model architecture}
\label{table:stack_architecture}
\centering
\begin{tabular}{lll} 
\toprule
Layer     & Output Shape & \# of Trainable Parameters  \\ 
\hline
Input     & 20           & 0                           \\
Embedding & (20, 8)      & 960000                      \\
LSTM      & (20, 256)    & 2055560                     \\
Dense     & (20, 256)    & 64416                       \\
Dense     & (10000)      & 970000                      \\
\hline
\end{tabular}
\end{table*}

\end{document}